\documentclass[journal]{IEEEtran}

\usepackage{cite}
\usepackage[pdftex]{graphicx}
\DeclareGraphicsExtensions{.pdf,.jpg} 

\usepackage{array}
\usepackage[caption=false,font=footnotesize]{subfig}
\usepackage{stfloats} 
\usepackage{url}
\usepackage[english]{babel}
\usepackage{blindtext}
\usepackage{bm,amssymb,amsmath,color,algorithm,algorithmic,url}

\begin{document}
\title{Optimal Deep Learning for Robot Touch}
\author{Nathan~F.~Lepora* and~John~Lloyd*
\thanks{This work was supported by an award from the Leverhulme Trust on `A biomimetic forebrain for robot touch' (RL-2016-39).}
\thanks{* NL and JL contributed equally to this work.}
\thanks{N. Lepora and J. Lloyd are with the Department
of Engineering Mathematics, Faculty of Engineering, University of Bristol and Bristol Robotics Laboratory, Bristol, UK. e-mail	{\tt\footnotesize \{n.lepora, jl15313\}@bristol.ac.uk}.}
}

\markboth{Accepted in the IEEE Robotics \& Automation Magazine special issue on Deep Learning and Machine Learning in Robotics}%
{N. Lepora and J. Lloyd}

\maketitle

\begin{abstract}
This article illustrates the application of deep learning to robot touch by considering a basic yet fundamental capability: estimating the relative pose of part of an object in contact with a tactile sensor. We begin by surveying deep learning applied to tactile robotics, focussing on optical tactile sensors, which help bridge from deep learning for vision to touch. We then show how deep learning can be used to train accurate pose models of 3D surfaces and edges that are insensitive to nuisance variables such as motion-dependent shear. This involves including representative motions as unlabelled perturbations of the training data and using Bayesian optimization of the network and training hyperparameters to find the most accurate models. Accurate estimation of pose from touch will enable robots to safely and precisely control their physical interactions, underlying a wide range of object exploration and manipulation tasks.
\end{abstract}


%
\IEEEpeerreviewmaketitle

\begin{figure*}[b!]
	\vspace{-2em}
	\centering
	\includegraphics[width=2\columnwidth,trim={30 90 30 90},clip]{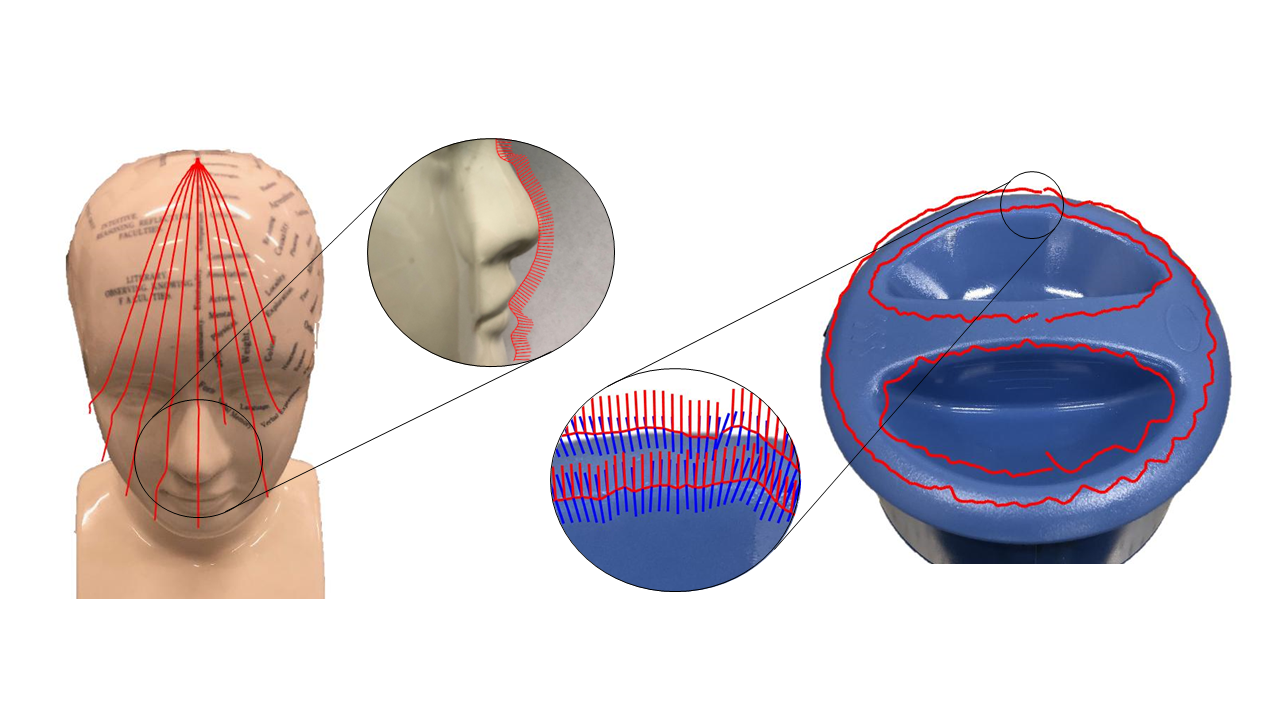} 
	\caption{Robot trajectories on a complex 3D surface and edge (a porcelain bust and container top) using pose estimation while sliding over the object.}
	\label{fig1}
\end{figure*}

\section{Introduction}

\IEEEPARstart{O}{ur} primary human senses of vision, audition and touch enable us to interact with a complex and ever-changing environment. Vision and audition are distal senses, with which we reason about and plan our interactions. In contrast, touch is a proximal sense that enables us to interact directly with our surroundings, either to avoid harm or to explore and manipulate nearby objects. It has become something of a cliche to remark that if we want robots to interact in a useful way with our world, then they will need versions of these three senses and the intelligence to use them effectively.  

However, there is a huge disparity between the research effort invested in applying modern artificial intelligence to distal senses compared with proximal senses. While for vision, there has been an explosion of interest to many tens of thousands of research papers, this has not been the case for touch, with a more gradual increase to about 50 published studies. Only a handful of labs worldwide routinely apply deep learning to tactile sensing and fewer still apply that expertise to control physically interactive robots.

One potential explanation for this disparity is the degree of synergy between artificial vision and deep learning: originally, convolutional neural networks (ConvNets) were inspired by the neuroanatomy and layered hierarchy from the retina through to sub-cortical then cortical visual processing structures of the brain, and the 2012 win of the ImageNet competition with \textcolor{black}{AlexNet sparked a revolution in computer vision}. However, deep learning has spread quickly to other applications such as speech recognition, so this initial impetus to vision does not explain the later disparity with touch. 

In our view, the main barriers to applying deep learning to touch are: (1) a lack of cheap, robust and easy-to-use artificial tactile sensors, contrasting with modern cameras for computer vision; (2)~the difficulty of obtaining high-quality tactile data due to a lack of public repositories and because a robot is usually needed to investigate the most interesting research questions; and (3) a lack of interest in the AI community on applying deep learning to touch. The latter barrier seems almost paradoxical when most reports on AI aimed at policy makers and the public are filled with pictures of humanoid robots that will be useless in practice without functional hands.

\textcolor{black}{In this article, we illustrate the application of deep learning to robot touch by considering a basic yet fundamental capability: estimating the relative pose of part of an object in contact with a fingertip. Tactile sensors can estimate the pose of the region of the object being contacted by inverting the tactile image into geometric features of the contact. However, finding the relation between high-dimensional tactile images and low-dimensional pose is a challenge: tactile sensors like our fingertips are soft and curved, so physical interaction deforms the sensor in complex ways depending on the object shape, contact forces and contact history. In our view, this difficulty has confined the use of robot touch to very primitive tasks compared with the fine motor capabilities of humans.}
	
\textcolor{black}{Accurate estimation of pose from touch will enable robots to safely and precisely control their physical interactions. For example, pose information can enable precise control of a fingertip sliding over complex objects, analogously to how humans trace our fingers over novel objects to explore shape. Previous work~\cite{Lepora2019a} demonstrated contour following around planar objects in 2D using deep learning applied directly to the tactile images, but required that the network be carefully hand-tuned otherwise the pose estimate would fail as the sensor sheared while sliding over the object. Here we adopt a different approach of collecting training data that simulates the effect of shear, and then use a black-box Bayesian optimizer to select the network architecture and other associated hyperparameters. In consequence, we demonstrate controlled sliding motion over complex 3D objects (Figure~\ref{fig1}).}

\begin{figure*}[t!]
	\centering
	\begin{tabular}{@{}c@{}c@{}c@{}c@{}c@{}c@{}c@{}c@{}}
		\multicolumn{4}{c}{\textbf{TacTip biomimetic optical tactile sensor}} & \multicolumn{4}{c}{\textbf{GelSight optical tactile sensor}} \\
		\multicolumn{4}{c}{\includegraphics[width=0.75\columnwidth]{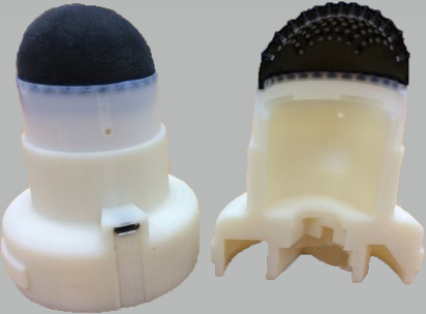}} & 
		\multicolumn{4}{c}{\includegraphics[width=0.75\columnwidth,trim={55 20 60 20},clip]{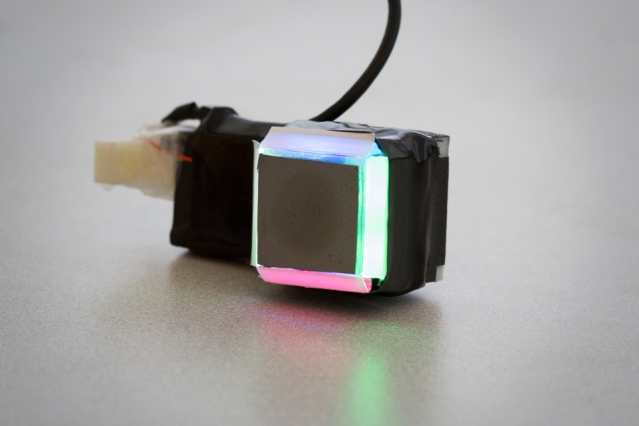}} \\
		\multicolumn{4}{c}{\textbf{TacTip tactile images}} & \multicolumn{4}{c}{\textbf{GelSight tactile images}} \\
		\includegraphics[width=0.25\columnwidth,trim={300 200 300 100},clip]{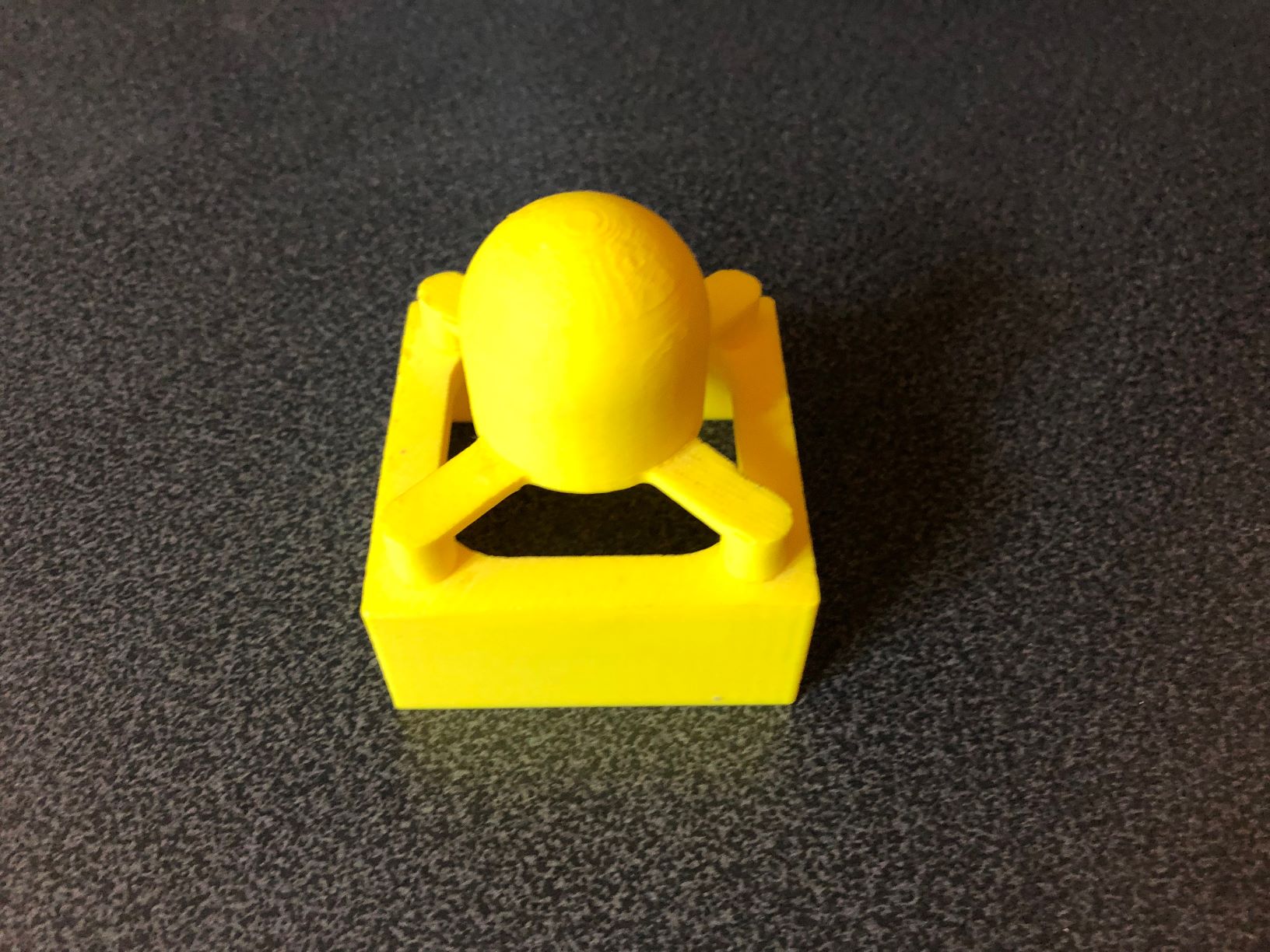} &
		\includegraphics[width=0.25\columnwidth,trim={300 200 300 100},clip]{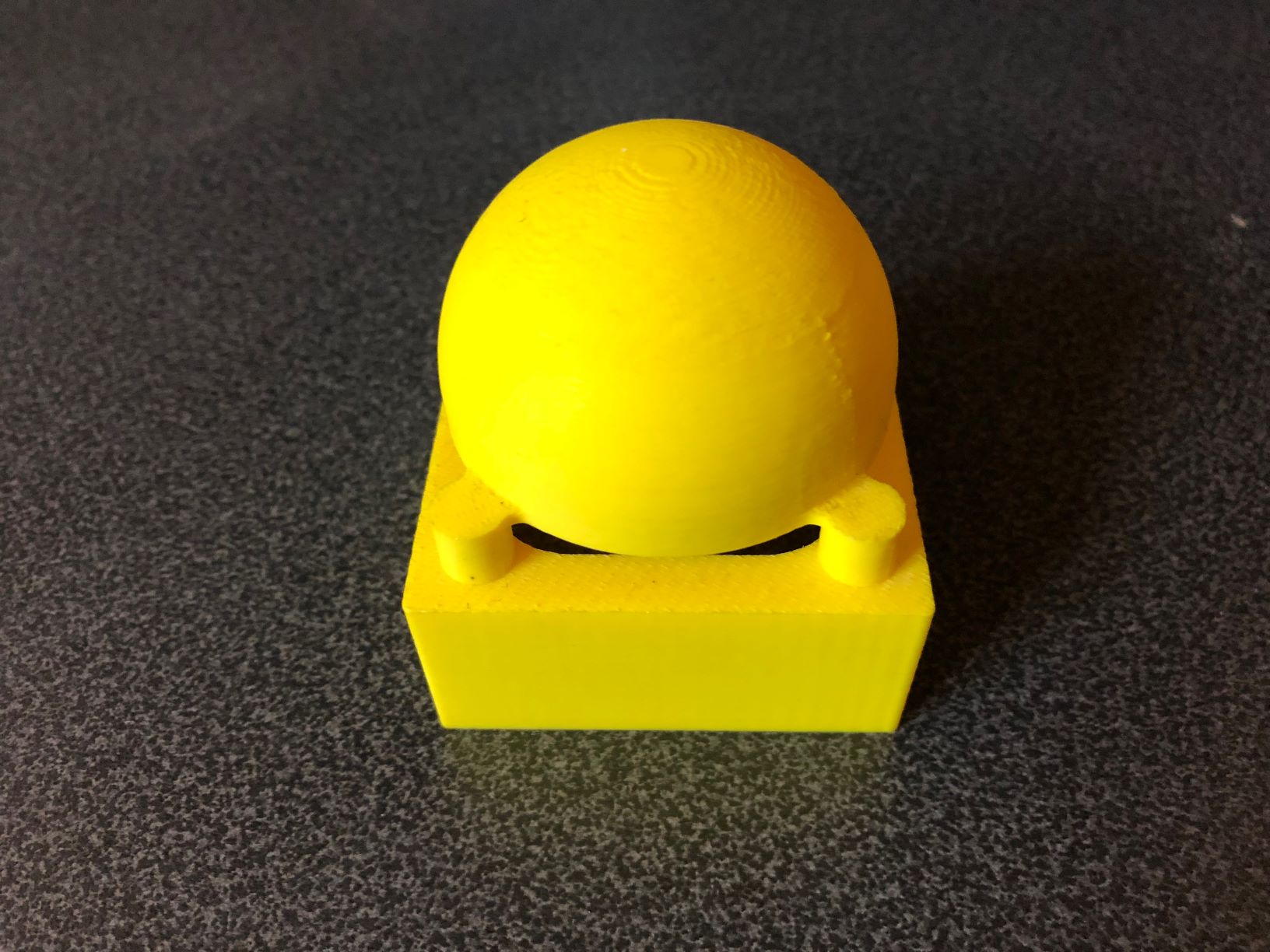} &
		\includegraphics[width=0.25\columnwidth,trim={300 200 300 100},clip]{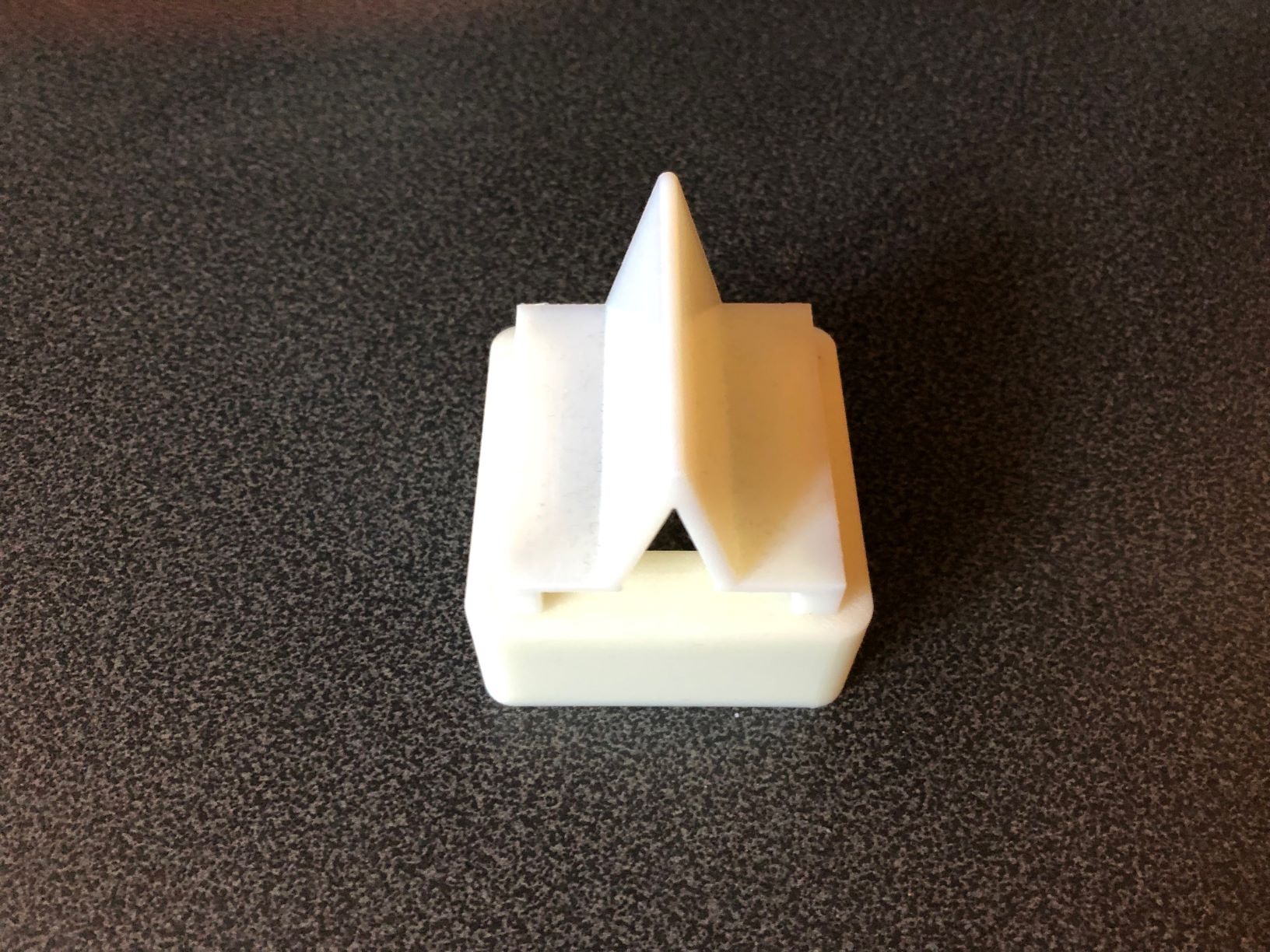} &
		\includegraphics[width=0.25\columnwidth,trim={300 200 300 100},clip]{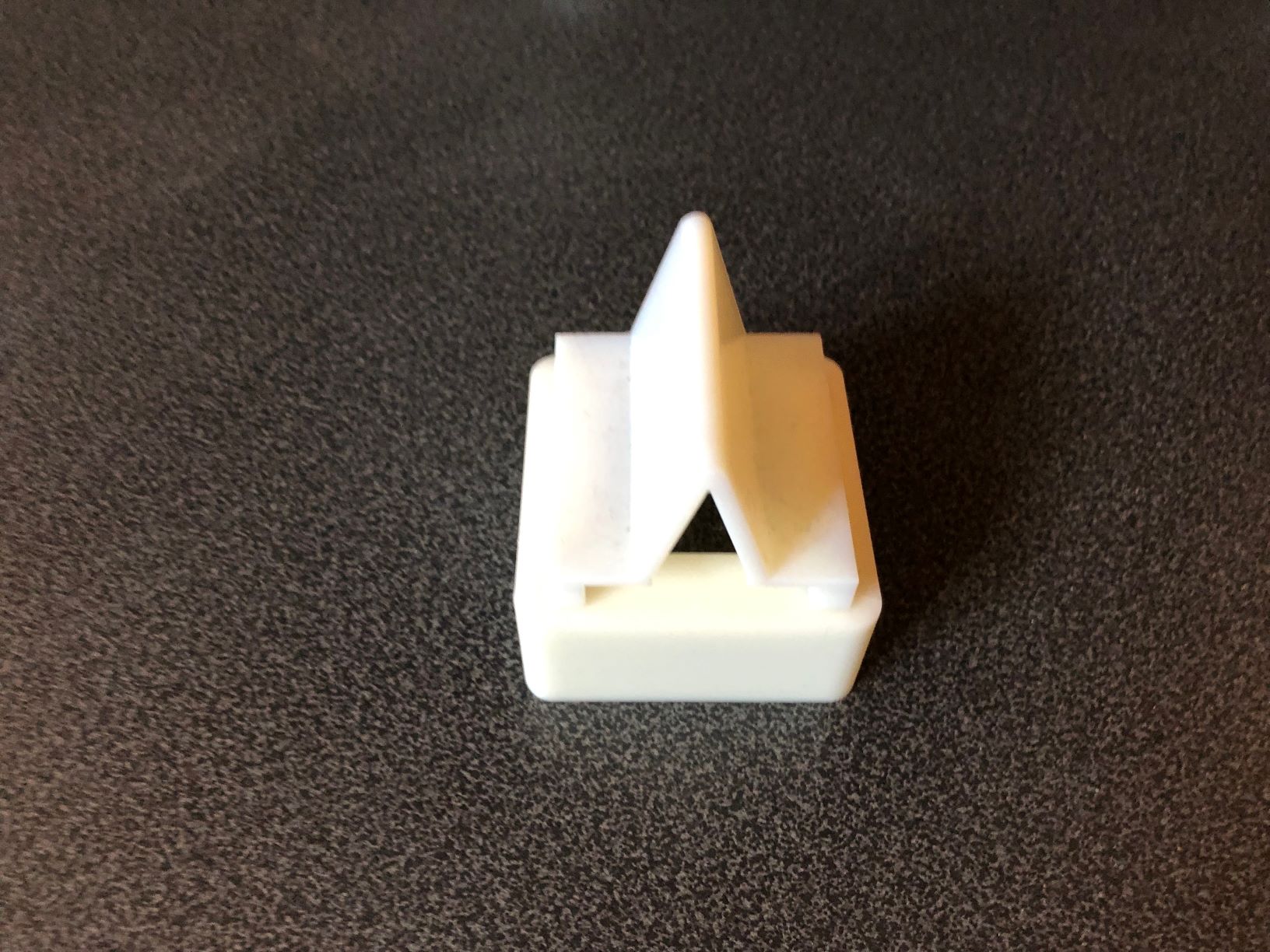}	&
		\multicolumn{4}{c}{\includegraphics[width=1\columnwidth,trim={0 -15 0 0},clip]{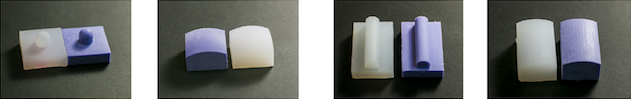}}\\ 
		\multicolumn{2}{c}{Hemisphere} & \multicolumn{2}{c}{Edge} & \multicolumn{2}{c}{\hspace{4em}Hemisphere} & \multicolumn{2}{c}{\hspace{2.5em}Cylinder} \\  
		\includegraphics[width=0.25\columnwidth,trim={100 0 100 0},clip]{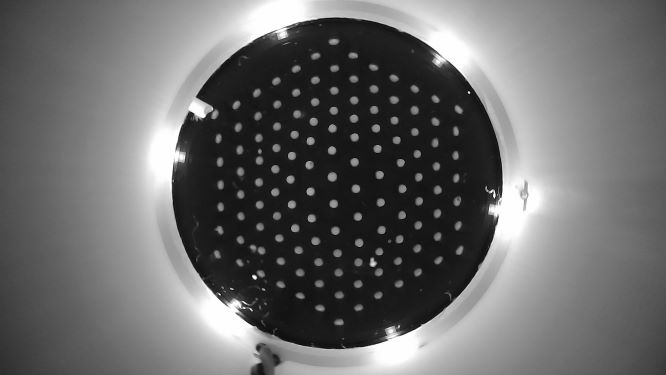} &
		\includegraphics[width=0.25\columnwidth,trim={100 0 100 0},clip]{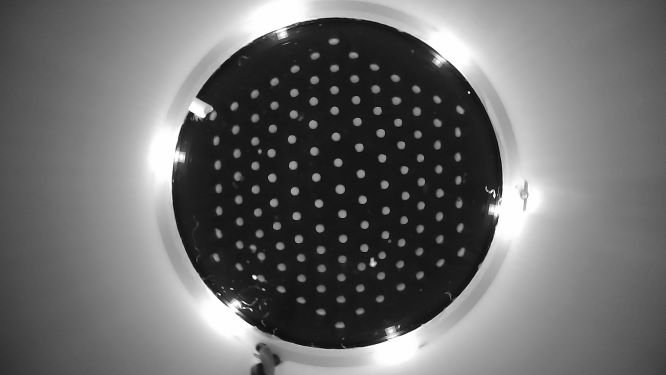} &
		\includegraphics[width=0.25\columnwidth,trim={100 0 100 0},clip]{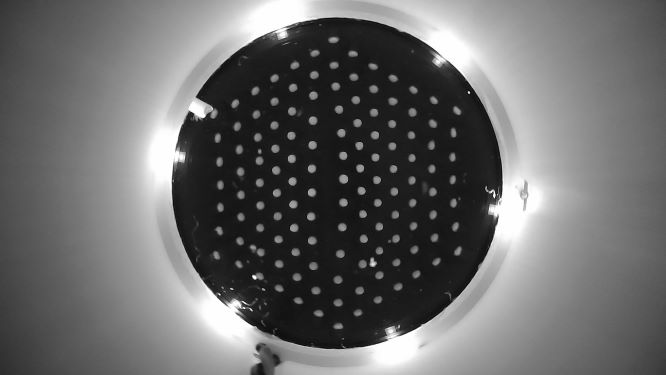} &
		\includegraphics[width=0.25\columnwidth,trim={100 0 100 0},clip]{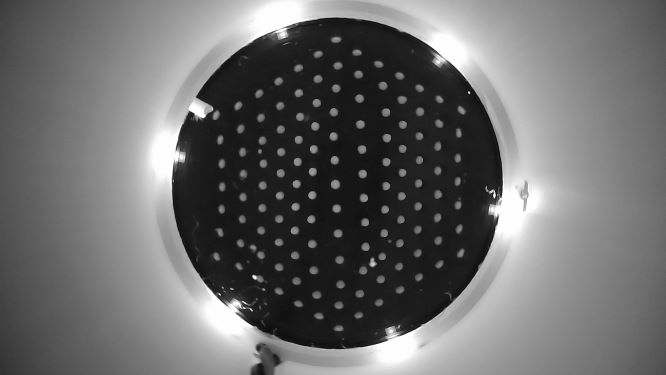} &
		\multicolumn{4}{c}{\includegraphics[width=1\columnwidth,trim={0 -5 0 0},clip]{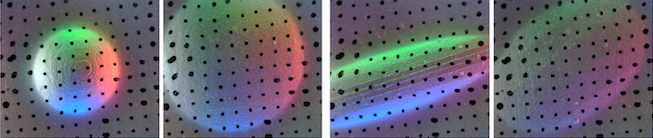}} \\ 
	\end{tabular}	
	\caption{The BRL tactile fingertip (TacTip) and GelSight optical tactile sensors. The bottom panels show examples of tactile images from manual contact against a few test stimuli. The GelSight tactile images are from~\cite{Yuan2017b} and the TacTip tactile images were generated for this paper.}
	\label{fig2}
	\vspace{-0em}
\end{figure*}

The main contributions of this research are to:
\begin{enumerate}
	\item Show how deep learning can be used to train accurate models to estimate 3D pose from tactile images.
	\item Develop pose estimation models that are insensitive to nuisance variables such as motion-dependent shear by directly incorporating these into the data collection.
	\item Introduce a systematic approach to model selection, which is needed for the most accurate models yet has not been used before for touch. 
\end{enumerate}
We also take the opportunity to survey deep learning applied to tactile robotics. In particular, we focus on optical tactile sensors (Figure~\ref{fig2}) that use an internal camera to image skin deformation, as these sensors help bridge from advances in deep learning for vision to the new domain of touch. 

\section{\textcolor{black}{Background}}

\subsection{Deep learning for tactile sensing}

Initial applications of deep learning to artificial tactile sensing were with taxel-based sensors, composed of discrete tactile elements embedded in a skin. The first study was in 2014, on tactile object recognition using deep learning and dropout~\cite{Schmitz2014}, which used a four-fingered robot hand with pressure-sensitive capacitive tactile arrays on the palm and finger joints/tips. Overall, they could recognize 20 objects held in a variety of grasps at about 90\% success rate using a ConvNet with 241 tactile inputs and 71 motor angles, currents and force/torque readings. They also observed that they were likely the first in this area because of the difficulty of gathering many samples of high-dimensional data with tactile sensors.

Over the next couple of years (2015-16), a handful of studies followed. These applied deep learning to pressure-sensitive tactile arrays on the fingertips of robot hands using ConvNets for tactile shape recognition~\cite{Cao2015}, material texture classification~\cite{Baishya2016} and slip identification~\cite{Meier}, and a recurrent LTSM network for classifying held objects~\cite{Buscher2015}. These studies used spatial and temporal data from their tactile arrays, which in combination gives a high dimensional input suitable for deep learning. Related approaches have continued since, with a variety of taxel-based sensors and hands.

A related area that combined tactile and visual data stemmed off from this early work in 2016. First applied to haptic adjectives from visual images and a single BioTac sensor~\cite{Gao2016}, the authors observed that neural networks serve as a natural unifying framework for multi-modal signal fusion. Soon after, visuo-tactile sensing was applied to object classification and grasp planning with an RGB-D camera and tactile arrays on a robot hand~\cite{Sun2016}. This area continued to develop since with multiple studies connecting look and feel~\cite{Yuan2017b,Calandra2018,Luo2018}, progressing more recently to deep learning methods that can transform between~\cite{Lee2019} or match visual and tactile data~\cite{Lin2019}.

\subsection{Optical tactile sensing for deep learning} \label{sec:4}

\begin{figure}[t]
	\centering
	\begin{tabular}[b]{@{}c@{}c@{}}
		\textbf{TacTip-enabled hand} & \textbf{GelSight-enabled gripper} \\
		\includegraphics[width=0.5\columnwidth,trim={0 25 225 0},clip]{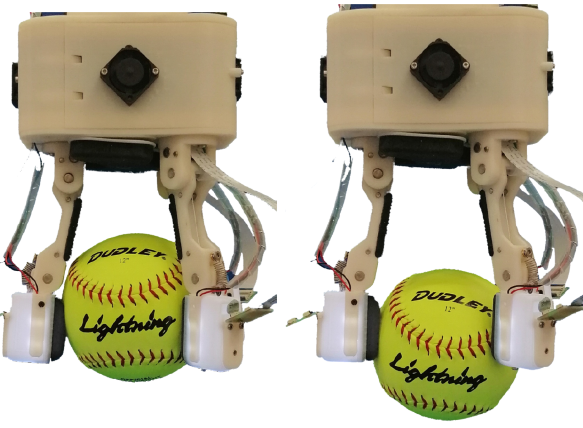} &
		\includegraphics[width=0.5\columnwidth,trim={0 50 0 0},clip]{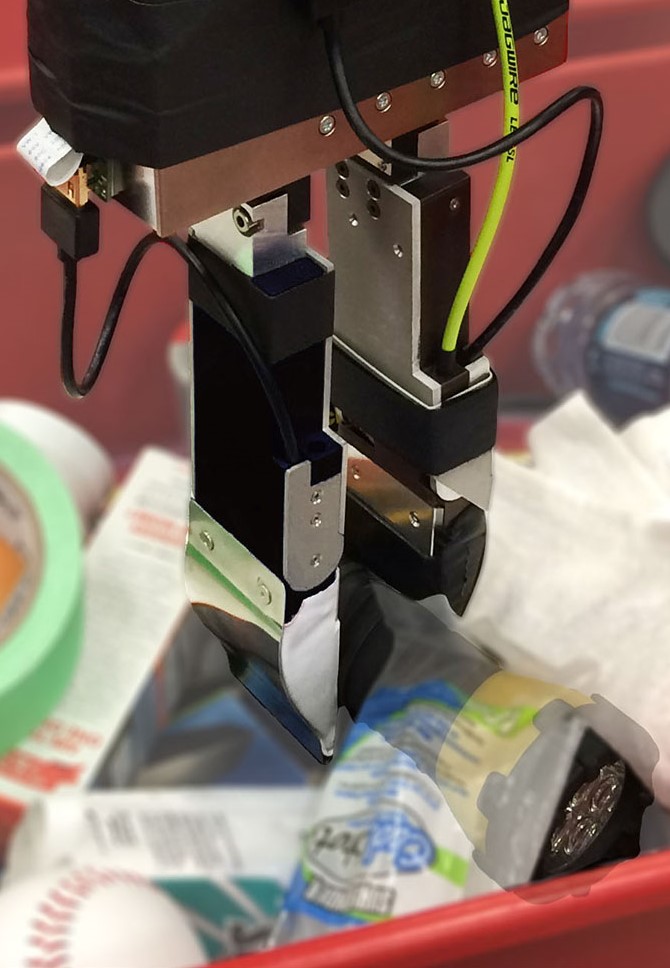} 
	\end{tabular}
	\caption{The TacTip and GelSight optical tactile sensors integrated with a robot hand and gripper. The TacTip-enabled hand is the BRL Tactile \mbox{Model-O} (T-MO)~\cite{Church2019}; the GelSight is integrated as the GelSlim sensor~\cite{Hogan2018}.}
	\label{fig3}
\end{figure}

Over the last few years (2017-19), the adoption of {\em optical} tactile sensors by several research groups has found a natural synergy with deep learning (Figures~\ref{fig2},\ref{fig3}). These optical tactile sensors use an internal camera to image the deformation of a compliant sensing surface in contact with a physical stimulus. Although optical methods have been considered promising for tactile sensing since the 1980s~\cite{Dario1984}, in practice the field had been dominated by various means of electromechanical transduction. More recently, however, a coherent body of research on deep learning for robot touch has used optical transduction to make progress on previously intractable problems, as typified by most of the recent progress in visuo-tactile sensing~\cite{Yuan2017b,Calandra2018,Luo2018,Lee2019,Lin2019}.

The first application of deep learning to an optical tactile sensor was for shape-independent hardness estimation using the GelSight~\cite{Yuan2017c} (Figure~\ref{fig2}, right). Tactile images (960$\times$720 pixels) were fed through a pre-trained deep ConvNet into a recurrent (LTSM) network to predict the Shore hardness of the contacted object, training with 7000 video sequences (5 frames each) over various object shapes and hardnesses. A major benefit was that the hardness estimation was insensitive to nuisance variables such as object shape and loading of the contact that have a complicated influence on the sensor output, which would be very difficult to model otherwise. 

Since that first study, the GelSight with deep learning has been used for various tactile robotic problems, including most of the visuo-tactile studies mentioned above~\cite{Yuan2017b,Calandra2018,Luo2018,Lee2019,Lin2019}, control of rolling motion~\cite{Tian2019}, and for grasp stability~\cite{Calandra2017} and grip readjustment~\cite{Hogan2018} on two-digit grippers (Figure~\ref{fig3}). 

Recently, deep learning has been found highly suited for another type of optical tactile sensor: the BRL tactile fingertip, known as the TacTip (Figure~\ref{fig2}). The first study considered robust edge perception for exploring objects with contour following~\cite{Lepora2019a}. Tactile images (128$\times$128 pixels) were fed into a ConvNet to predict 2D edge pose, training with 2000 videos (5 frames each) over a range of edge angles and positions. The challenge was to make the pose estimation insensitive to motion-dependent shear during sliding, which was addressed by hand-tuning the network architecture to have a stack of input convolutional layers that learned broader features across the tactile image.

The BRL TacTip has since been used with deep learning to identify benchmark objects and predict grasp success on a 3-finger robot hand~\cite{Church2019} (Figure~\ref{fig3}, right), and to extract features for predicting shape, 2D edge pose and force with a convolutional autoencoder~\cite{Polic2019}. These studies used standard ConvNet architectures to make predictions that are insensitive to nuisance variables such as the unknown object pose after automated grasping.

\subsection{Why optical tactile sensing?}

Given our opening comments about the barriers to applying deep learning to touch, why has a large proportion of research in that area used optical tactile sensing? In our view, the GelSight and TacTip tactile sensors share some commonalities that suit deep learning; for example, both use an internal camera and internal light sources to image the skin deformation. That said, there are also significant differences, which have implications for their future development and use.

Historically, both sensors were invented before the present revolution in AI: the first paper on the GelSight was in 2009, on retrographic sensing for the measurement of surface texture and shape~\cite{Johnson2009}; likewise, the first paper on the BRL TacTip was also in 2009, on the development of a tactile fingertip based on biologically-inspired edge encoding~\cite{Chorley2009}. Since then, both sensors have progressed in their designs and uses, but their operating principles have remained the same~\cite{Yuan2017,Ward-Cherrier2018}.

The GelSight operates on the principle of measuring normal indentation of a surface based on the shading of light reflected off the underside of that surface~\cite{Johnson2009,Yuan2017} so that, with multiple light sources, a depth map can then be reconstructed using a photometric stereo algorithm. The most common design uses internal RGB lights arranged symmetrically inside the sensor to give three sources of shading (Figure~\ref{fig2}, right). Although the original intention was to measure indentation directly, when used with deep learning the unprocessed RGB images are fed into the input layer of a ConvNet to extract tactile image features.

The BRL TacTip operates by measuring deformation of a surface from the shear displacement of markers on the tips of 3D pins protruding beneath that surface~\cite{Chorley2009,Ward-Cherrier2018} (Figure~\ref{fig2}, right); this design is biomimetic because human fingertips have an analogous structure in which mechanoreceptors lie on papillae protruding between the epidermal and dermal skin layers. Originally, the pin tips were detected and tracked to give tactile features, most commonly as a time series of $(x,y)$-displacements of the markers. However, when used with deep learning, the images are fed directly as inputs to a ConvNet. 

In our view, the GelSight and TacTip sensors are well suited for deep learning because they both produce tactile images where every pixel gives information about surface deformation. For the GelSight, every pixel has RGB intensity readings that relate to indentation. For the TacTip, every pixel can signal whether a pin has moved to that location or not. Moreover, in recent years, the GelSight design has been modified to have markers inside the skin to indicate shear~\cite{Yuan2017}, which is similar to the operation of the TacTip (but without its 3D-pin structure). Meanwhile, depth maps can be inferred indirectly from the TacTip pin positions (for example, by using a Voronoi transformation). 

\begin{figure*}[t]
	\centering
	\begin{tabular}[b]{@{}c@{}}
		\textbf{PoseNet convolutional neural network architecture} \\
		\includegraphics[width=2\columnwidth,trim={120 200 100 140},clip]{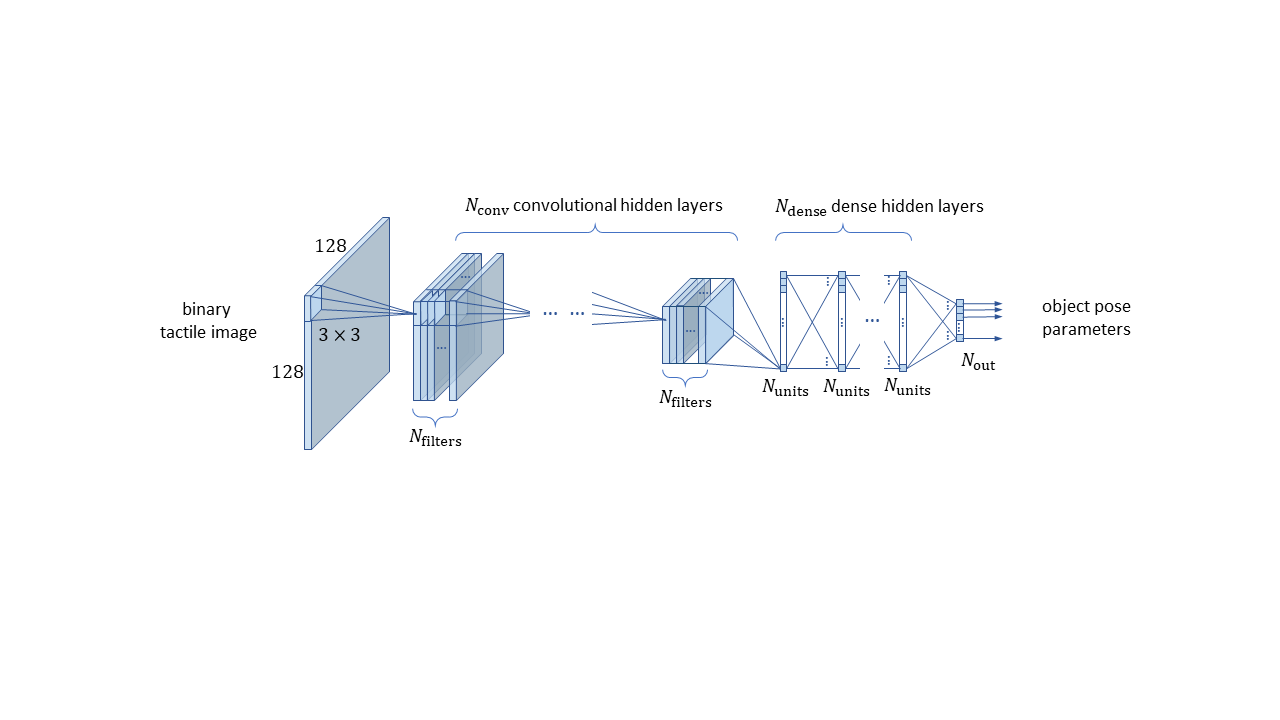} 
	\end{tabular}
	\vspace{0.5em}
	\caption{poseNet convolutional neural network. A binary sensor image is processed by a sequence of $N_{\rm conv}$ convolutional layers, each containing $N_{\rm filters}$ filters. The resulting convolutional features then pass through a sequence of $N_{\rm dense}$ fully connected hidden layers, each containing $N_{\rm units}$ processing units, to produce a high-level feature vector. The high-level features are then combined by a linear output layer to give $N_{\rm out}$ continuous-valued object pose parameters.}
	\label{fig4}
	\vspace{0.5em}
\end{figure*}

\section{PoseNet: 3D-pose estimation from touch}\label{sec:3b}

To illustrate the application of deep learning to robot touch, we considered local pose estimation of a 3D surface in contact with the BRL TacTip optical tactile sensor. Pose estimation is a fundamental capability of tactile sensing, because knowledge of relative pose enables a tactile robot to control its interactions. In the following, we refer to this neural network as a {\em PoseNet}, because it estimates 3D pose from a tactile image. 

Following previous work with optical tactile sensors~\cite{Lepora2019a}, we used a standard ConvNet architecture to generate and process tactile features to predict 3D pose (Figure~\ref{fig4}). The tactile image feeds forward through a sequence of $N_{\rm conv}$ convolutional and max-pooling hidden layers, using $N_{\rm filters}$ filters in each layer and $3\times3$ kernels with unit stride (no padding). The resulting tactile features then pass through a further $N_{\rm dense}$ fully-connected hidden layers, each with $N_{\rm units}$ processing units that compute a dense matrix multiplication plus bias. These processed features are combined linearly at the output layer to give $N_{\rm out}$ pose components.

Unlike previous work in deep learning for tactile sensing, we did not hand-tune these network parameters or rely on values from other studies. Instead, we treated them as unknown hyperparameters to be optimized in the training process. This approach has been used widely elsewhere in deep learning, but to the best of our knowledge this is the first time it was applied to touch.  

Other hyperparameters of the poseNet architecture and training process that were optimized include the activation function used in the hidden layers (ReLU or ELU) and regularization parameters to avoid overfitting the training data, encompassing batch normalization, dropout and L1/L2-regularization. Where batch normalization was used, it was inserted between the convolution operation and activation function in all convolutional hidden layers. Where dropout was used, it was applied to the inputs of all dense hidden layers and the output layer, before the dense matrix multiplication. Where used, L1- or L2-regularization were applied to the weights in all dense hidden and output layers. The reason for including more than one type of regularization approach is that they tend to work in complementary ways: batch normalization and dropout inject noise into the training process, helping to prevent complex co-adaptation of features, whereas L2-regularization encourages smooth mappings, and L1-regularization encourages sparse models.

In this work, we considered two distinct types of poseNet: one for a 3D surface and another for a 3D edge (Figure~\ref{fig4}). In addition to differences in their trained parameters, the networks have different outputs. The network for the 3D surface has 3 outputs: the depth, roll and pitch of the contact, and the network for the 3D edge has 5 outputs: horizontal distance, depth, roll, pitch and yaw. These parameters are sufficient to describe the pose of a tactile sensor in contact with a surface or contour, by defining the local pose relative to a plane or line on a tangent to those objects. 

\begin{figure}[b]
	\centering
	\begin{tabular}[b]{@{}c@{}c@{}}
		\textbf{3D surface} & \textbf{3D edge} \\
		\textbf{(3 pose parameters)} & \textbf{(5 pose parameters)} \\
		\includegraphics[width=0.55\columnwidth,trim={250 150 190 50},clip]{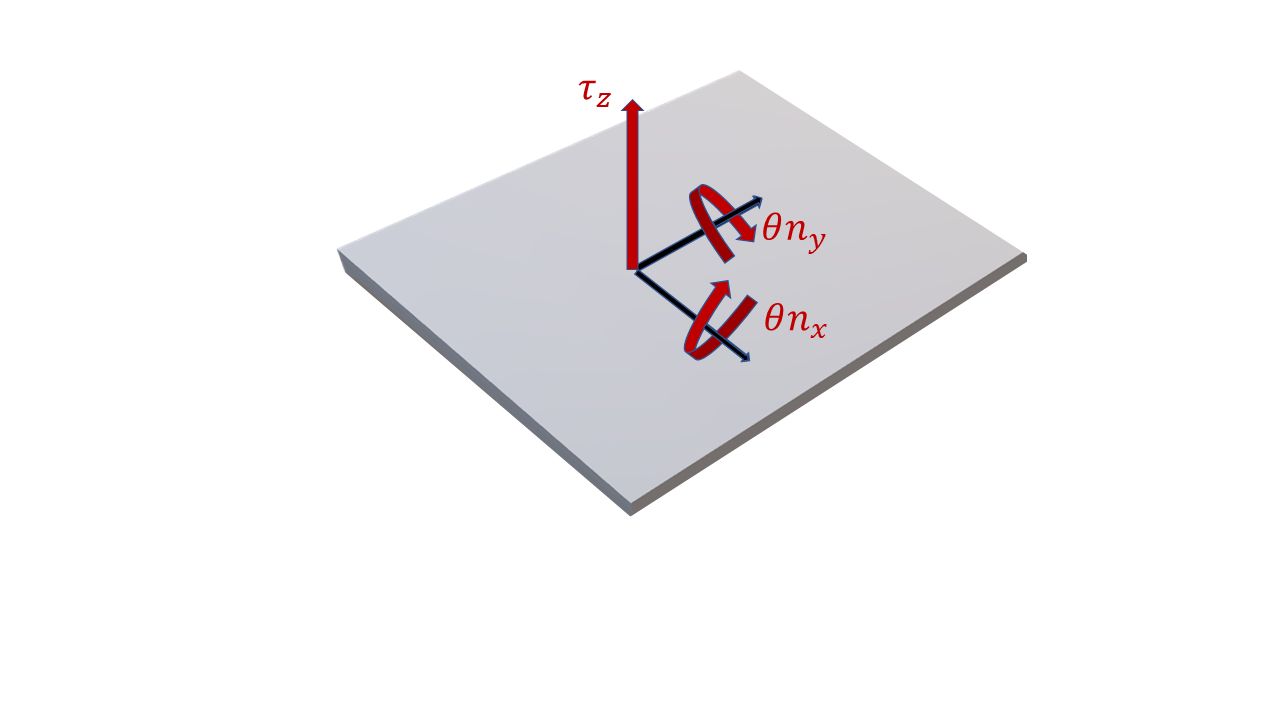} &
		\includegraphics[width=0.45\columnwidth,trim={300 190 300 30},clip]{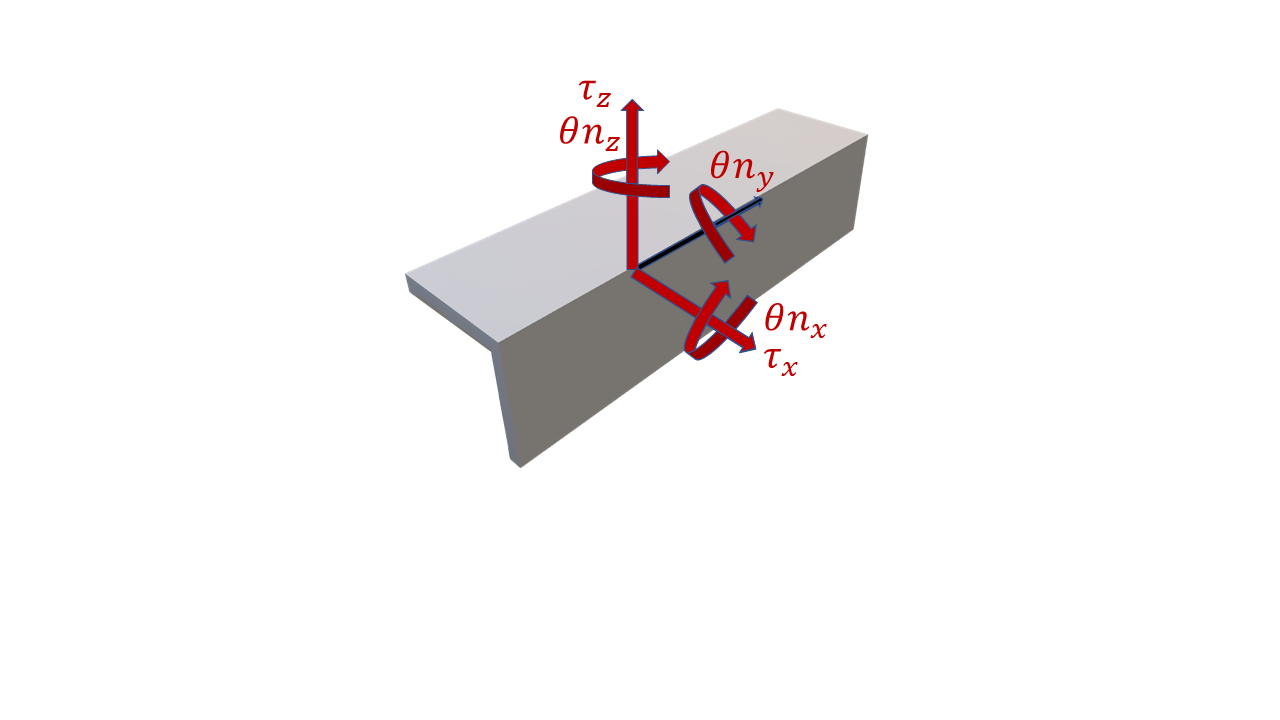} 
	\end{tabular}
	\caption{Pose parameters for the 3D surface and edge. The surface has 3 parameters: depth, roll and pitch; the edge has 5 parameters: $x$-horizontal, depth, roll, pitch and yaw. The labels are in axis-angle coordinates $(\bm\tau,\theta\bm n)$.}
	\label{fig5}
\end{figure}

\section{Training data and network optimization}

A contribution of this paper is to identify some subtleties in the data collection and hyperparameter optimization process when developing deep neural networks for robot touch. Robot touch senses via contact, so the manner of contact affects the tactile data. To predict pose accurately, the trained network should be insensitive to how the sensor has reached its current pose; for example, the motion of the sensor sliding across a surface or along an edge (Figure~\ref{fig1}). Here we included representative examples of these motions as unlabelled perturbations of the training data. However, these perturbations make it more difficult to predict pose, to the extent that improperly tuned hyperparameters gave extremely sub-optimal results. For a systematic approach, we used automatic tuning methods, specifically Bayesian optimization.  


\subsection{Training data collection}
\label{sec4a}

\begin{table}[b]
	\begin{tabular}{l|cc|c}
		\textbf{Parameter} & 
		\multicolumn{2}{c|}{\begin{tabular}{c}\textbf{Labelled}\\ \begin{tabular}{cc} \textbf{3D surface} & \textbf{3D edge}\end{tabular}\end{tabular}} &
		\begin{tabular}{c}\textbf{Unlabelled}\\ \textbf{Perturbation}\end{tabular} \\
		\hline
		$x$-horizontal & - & $[-5,5]\,$mm  & $[-5,5]\,$mm \\
		$y$-horizontal & - & -  & $[-5,5]\,$mm \\
		depth & $[-5,-1]\,$mm & $[-5,-1]\,$mm & $0\,$mm \\
	    roll & $[-15,15]\,$deg & $[-15,15]\,$deg & $[-5,5]\,$deg \\
		pitch & $[-15,15]\,$deg & $[-15,15]\,$deg & $[-5,5]\,$deg \\
		yaw & - & $[-45,45]\,$deg & $[-5,5]\,$deg \\
	\end{tabular}\\
	\caption{Training data pose parameter ranges.}
	\label{tab1}
\end{table}

\begin{table}[b]
	\begin{tabular}{l@{}c@{}c}
		\textbf{Parameter} & \textbf{Range} & \textbf{Distribution} \\ 
		\hline
		\# convolutional hidden layers, $N_{\rm conv}$ & $\{1, 2, 3, 4, 5\}$ & Uniform \\ 
		\# convolutional filters, $N_{\rm filters} $ & $\{2, 4, 8, \ldots, 512\}$ & Uniform \\ 
		\# dense hidden layers, $N_{\rm dense}$ & $\{1, 2, 3, 4, 5\}$ & Uniform \\
		\# dense hidden layer units, $N_{\rm unit}$ & $\{2, 4, 8, \ldots, 512\}$ & Uniform \\ 
		hidden layer activation function & \{ReLU, ELU\} & Uniform \\ 
		L1-regularization coefficient & $[10^{-4},10^{-1}]$ & Log uniform \\ 
		L2-regularization coefficient & $[10^{-4},10^{-1}]$ & Log uniform \\ 
		dropout coefficient & $[0, 0.5]$ & Uniform \\ 
		batch size & $\{16,32,64,128\}$ & Uniform \\ 
	\end{tabular}
	\caption{Search ranges used to optimize poseNet hyperparameters.}
	\label{tab2}
\end{table}

\begin{figure}[t]
	\centering
	\begin{tabular}{@{}c@{}}
		\includegraphics[width=\columnwidth,trim={62 45 40 25},clip]{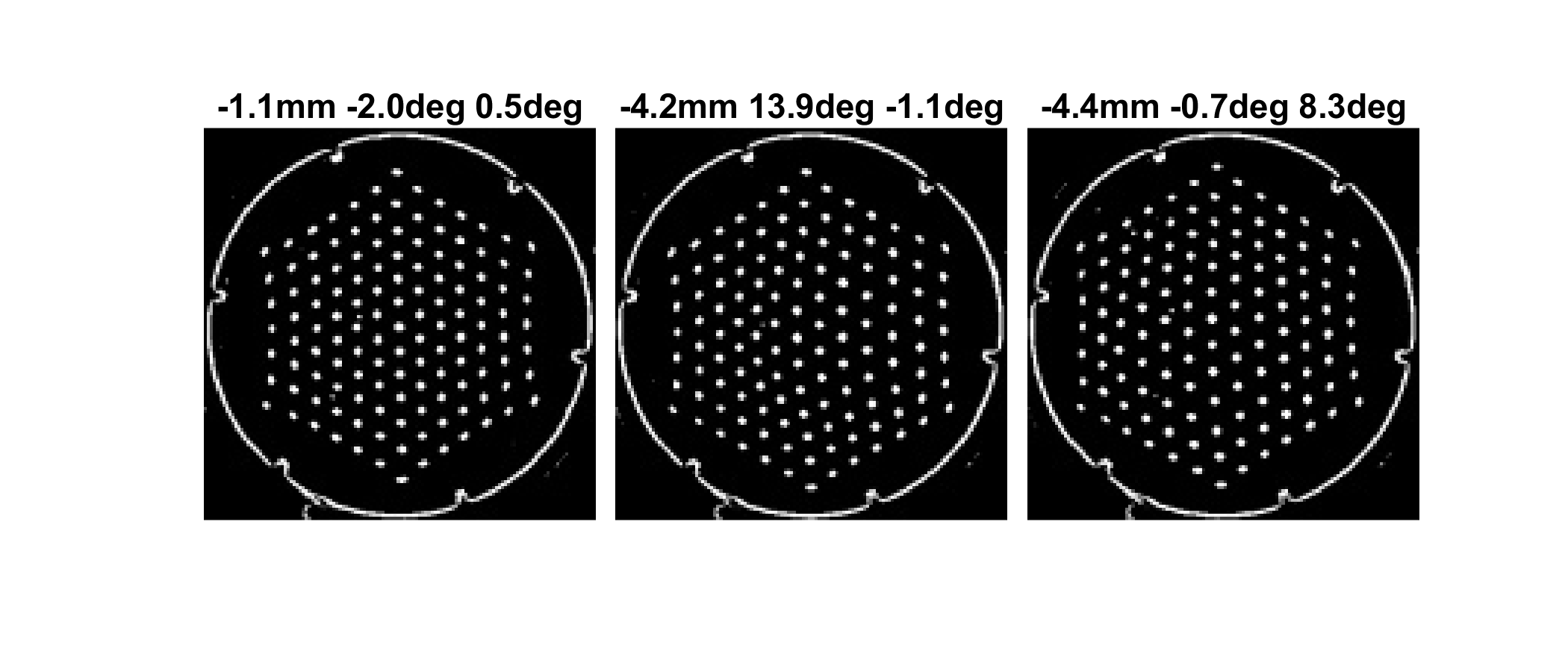} \\
		\includegraphics[width=\columnwidth,trim={62 45 40 25},clip]{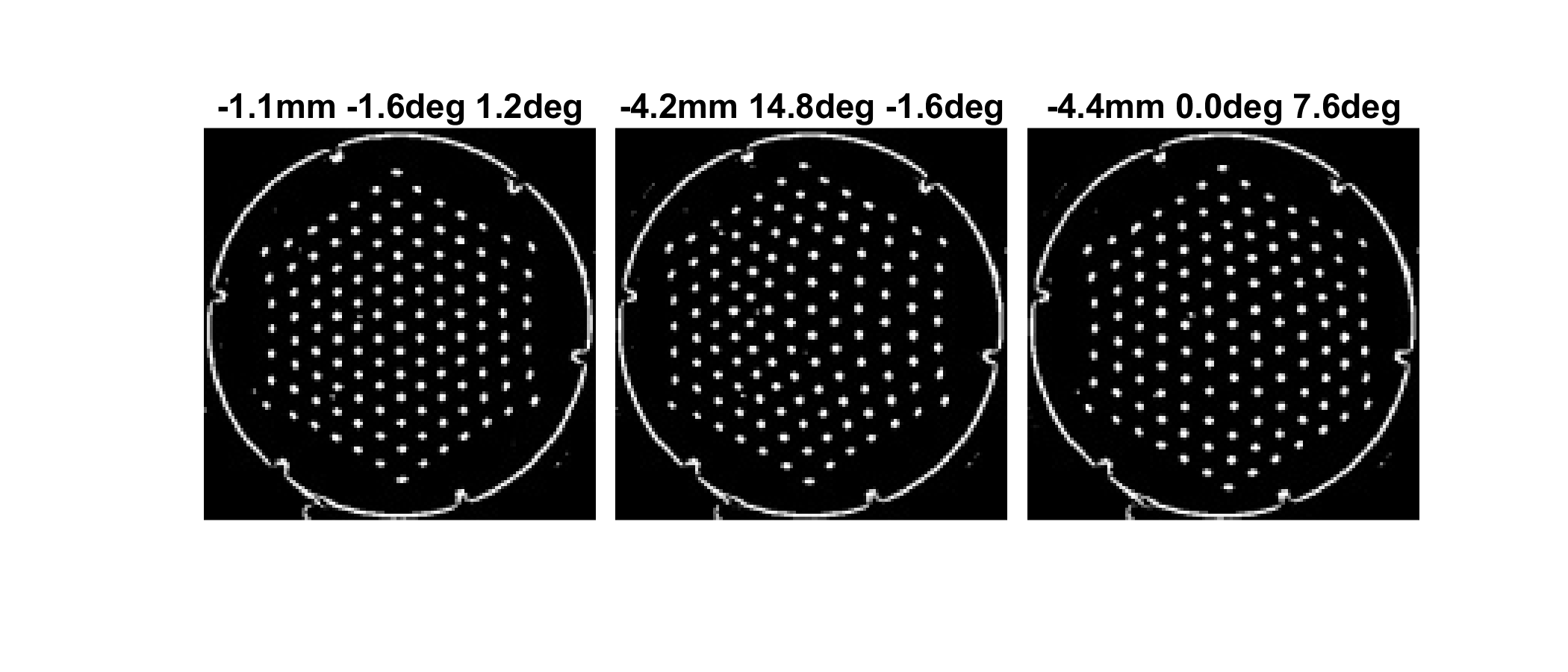} \\
	\end{tabular}
	\caption{Example pre-processed tactile images for the 3D surface and their pose labels (depth, roll and pitch). The tactile images were selected to have similar labels for each column (equal depth; roll and pitch within $1\,$deg). \textcolor{black}{Differences between rows are due to the unlabelled random perturbations in pose, and are most apparent for the strongest/deepest contacts} (right column).}
	\label{fig6}
	\vspace{-0.5em}
\end{figure}

Three data sets were collected \textcolor{black}{for each of the two objects considered here (a flat surface and straight edge; Figure~\ref{fig5})}, giving independent training, validation and testing data. Each set was over 2000 contacts with poses sampled randomly from uniform distributions within their allowed ranges (Table~\ref{tab1}). Separate training and validation sets were used to mitigate potential dataset shift, rather than using a randomized split. Data comprised single tactile images cropped and subsampled to a $128\times 128$ pixel region that views all tactile elements of the sensor surface. These images were then pre-processed with an adaptive threshold to produce binary (black/white) inputs to the neural network (examples in Figure~\ref{fig6}), \textcolor{black}{reducing sensitivity to changes of lighting inside the sensor.}

We found it crucial to mimic the effect of motion on the tactile data while collecting the training data, because motion-dependent shear affects measurements from optical tactile sensors such as the BRL TacTip. \textcolor{black}{Each sample of data has a random labelled pose and a random unlabelled shear perturbation (ranges in Table~\ref{tab1}). First the sensor is brought into contact at the labelled target pose minus the unlabelled perturbation, then moved to the target pose, then the tactile image collected.} The perturbations took uniform random values within ranges typical of movements that may move the sensor without damaging it, in all pose components except the vertical. Ranges for the target poses and the unlabelled perturbations were chosen according to how much the sensor can move to remain safely in contact without causing damage.

\subsection{Model training}

As is common practice in neural network regression problems, the network weights and biases are optimized during training by minimizing the mean-squared error (MSE) between the predicted outputs and target labels. The loss components were weighted by 1/maximum$^2$ for the parameters in Table~\ref{tab1} ({\em e.g.} depth 1/5$^2$, roll 1/15$^2$, yaw 1/45$^2$) to give peak losses of order one. Several different regularization approaches were used to avoid overfitting the training data, including constraining the number and size of hidden layers, and use of early stopping during training. These choices were part of the hyperparameter optimization process describe below. 

All PoseNet models were trained using the Adam adaptive learning rate optimizer (initial learning rate $10^{-4}$; decay coefficient $10^{-6}$). This optimizer consistently gave good solutions and seems to converges more quickly than other types of optimizer. Prior to training, all weights were initialized to small random values. Early stopping was used to terminate the optimization process, where the MSE loss was computed for a separate validation data set and the process halted when there was no further improvement over 10 epochs.

Training and optimization of the deep neural networks was implemented in the Keras library on a Titan Xp GPU \textcolor{black}{(12Gb memory)} hosted on a Windows 10~PC. A training run typically takes 5 to 20 minutes depending on the size of the network. 

\subsection{Hyperparameter optimization}

\begin{figure*}[t]
	\centering
	\begin{tabular}[b]{@{}c@{}}
		\textbf{PoseNet predictions (3D surface)} \\
		\includegraphics[width=2\columnwidth,trim={-50 0 -50 12},clip]{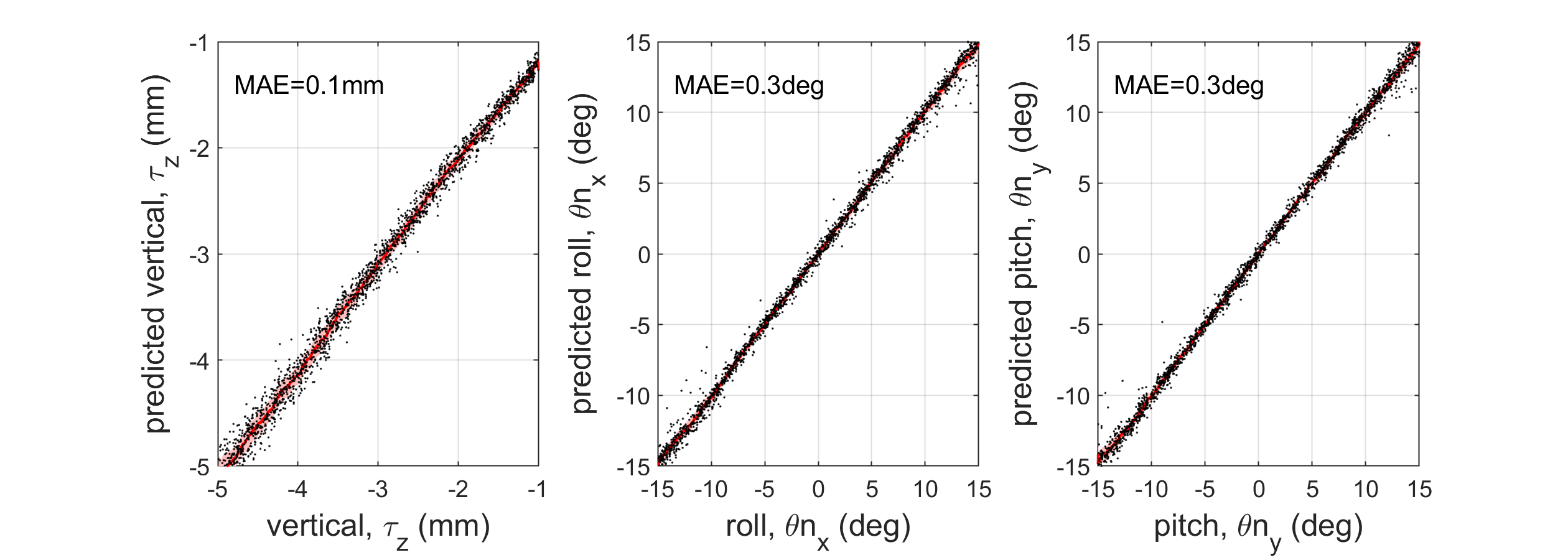} \\
	\end{tabular}
	\caption{Optimal PoseNet performance for the 3D surface. The smoothed predictions (red; 100-sample moving average), region within the smoothed absolute error (pink) and mean average errors (MAE) are shown.}
	\label{fig7}
\end{figure*}

Bayesian optimization was used to give a systematic approach to setting the network and training hyperparameters. The PoseNet models were optimized using 300 cost function evaluations, using 50 start-up random evaluations, from the MSE loss on the validation data set after early stopping.

\begin{figure*}[t!]
	\centering
	\begin{tabular}[b]{@{}c@{}}
		\textbf{\textcolor{black}{Hyperparameter optimization (3D surface)}} \\
		\includegraphics[width=2\columnwidth,trim={95 35 95 35},clip]{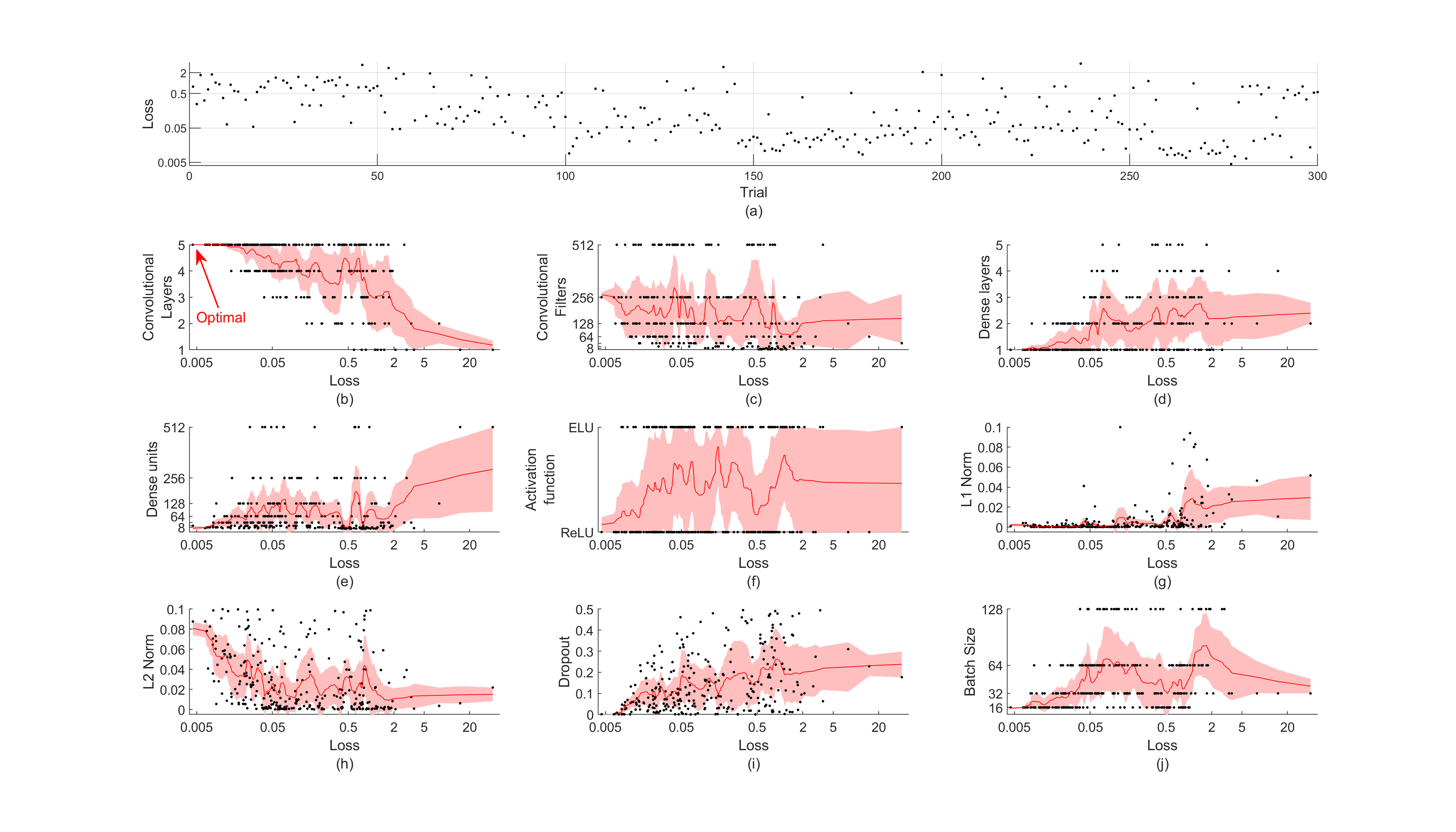}
	\end{tabular}
	\caption{\textcolor{black}{Hyperparameter optimization for the 3D surface. Top left panel: loss against trial number. Other panels: hyperparameter dependence on loss; optimal network to left of scatter plot.}}
	\label{fig8}
\end{figure*}

Bayesian optimization is a sequential, derivative-free, black-box optimization method that is well-suited to noisy and expensive-to-evaluate cost functions. It incrementally constructs an acquisition function that directs cost function sampling to areas where improvement is likely. We used a Python implementation of Bayesian optimization called HyperOpt (hyperopt.github.io/hyperopt), which constructs a generative, non-parametric model of an acquisition function called a Tree-structured Parzen Estimator (TPE). Overall, the optimization and training processes took 1-2 days on our hardware.

The model configurations were constrained to lie within a search space bounded by parameter ranges that take account of the available computing resources and reasonable values of the hyperparameters~(Table~\ref{tab2}). For example, given the $128\times128$-dimension of the input tactile image and the $2\times2$ max-pooling architecture of the network, the number of convolutional layers $N_{\rm conv}\leq 5$, and computing resources constrained the number of filters per layer $N_{\rm filter}\leq 512$ and \textcolor{black}{batch size $\leq 256$}. We emphasize that it was not obvious {\em a priori} where the optimal parameter values would be, so these ranges are the broadest that were reasonable for the problem; other network and learning hyperparameters could be considered, but this set of 10 parameters seemed reasonable based on past experience with deep learning using this sensor. 

\section{Tactile pose estimation for a 3D surface}
\label{sec:5}

\subsection{3D surface model results}

The performance of the trained poseNet was assessed on a third test dataset gathered for this purpose, comprising 2000 tactile images labelled with depth, roll and pitch pose parameters. The data were subject to random unlabelled perturbations in the target pose prior to collecting the tactile image, \textcolor{black}{using the procedure described in Section~\ref{sec4a}}. All labelled poses and unlabelled perturbations were sampled randomly within the same ranges as the training and validation data (Table~\ref{tab1}).

Highly accurate performance was obtained for the optimized PoseNet, with a close match between the predicted and labelled pose parameters (Figure~\ref{fig7}). The mean absolute error (MAE) between predictions and labels was used to summarize model performance, giving values $0.1\,$mm for depth, and $0.3\,$deg for roll and pitch. These results are precise compared with the size of the tactile sensor ($40\,$mm dia. tip) and pin spacing ($4\,$mm); also, this TacTip was not designed for this task, but is a standard version used in our lab~\cite{Ward-Cherrier2018}. \textcolor{black}{We emphasise that the precise accuracies hold even though the data were randomly perturbed by an unknown shearing motion that majorly affects the tactile image (Figure~\ref{fig6}).}

\subsection{Analysis of 3D surface model optimization}

\begin{table}[b]
	\begin{tabular}{l@{}cc}
		\textbf{Parameter} & \textbf{3D surface} & \textbf{3D edge} \\ 
		\hline
		\# convolutional hidden layers, $N_{\rm conv}$ & 5 & 5 \\ 
		\# convolutional kernels, $N_{\rm filters} $ & 512 & 256 \\ 
		\# dense hidden layers, $N_{\rm dense}$ & 1 & 2 \\
		\# dense hidden layer units, $N_{\rm unit}$ & 16 & 512 \\ 
		hidden layer activation function & ReLU & ReLU \\ 
		dropout coefficient & 0.001 & 0.203 \\ 
		L1-regularization coefficient & 0.001 & 0.0001 \\ 
		L2-regularization coefficient & 0.064 & 0.0003 \\ 
		batch size & 32 & 16 \\ 
	\end{tabular}
	\caption{Optimal PoseNet hyperparameters.}
	\label{tab3}
\end{table}

The benefit of optimizing the hyperparameters is evident from the 1000-fold decrease in validation loss over the 300 trials of training (Figure~\ref{fig8}, top-left panel). 

The first 50 trials were a period of high, scattered loss during the start-up evaluations ($0.05\lesssim{\rm MSE}\lesssim2$). Then within another 20 trials, the optimizer quickly found a good model (${\rm MSE}\simeq0.01$). Another 150 trials were needed before the optimizer improved substantially on this to give the best loss (${\rm MSE}~\simeq~0.005$), by which time it consistently gave models with low losses. 

\begin{figure*}[t]
	\centering
	\begin{tabular}[b]{@{}c@{}}
		\textbf{PoseNet predictions (3D edge)} \\
		\includegraphics[width=2\columnwidth,trim={100 0 90 0},clip]{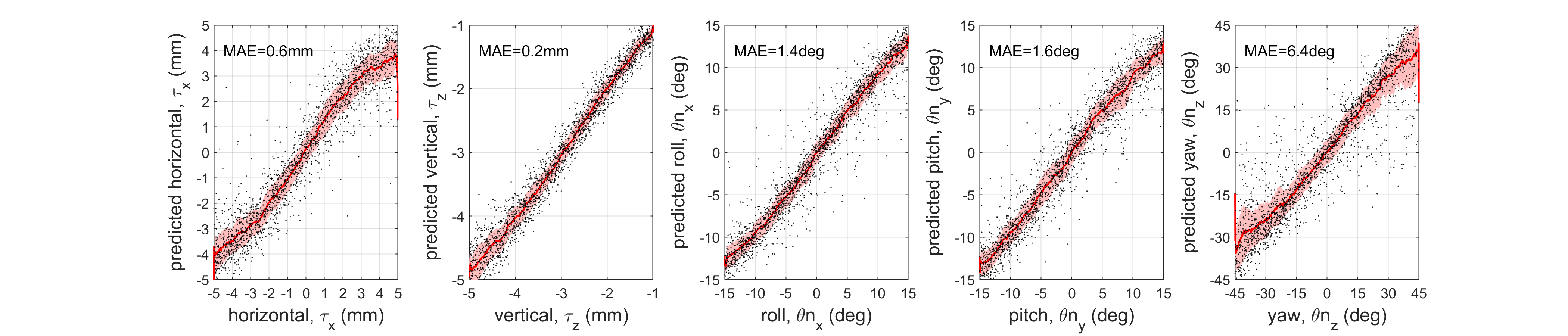} \\
	\end{tabular}
	\caption{Optimal PoseNet performance for the 3D edge. The smoothed predictions (red; 100-sample moving average), region within the smoothed absolute error (pink) and mean average errors (MAE) are shown. }
	\label{fig9}
\end{figure*}

The convergence of the optimization process can also be seen by plotting the hyperparameters against their corresponding losses, visualized as scatter plots ordered by loss (Figure~\ref{fig8}). As the loss becomes small (to the left), the hyperparameter values become more consistent near the optimum rather than being scattered. 

Overall, the optimization favoured network architectures~(Table~\ref{tab3}) that have a deep convolutional stage (5~hidden layers), a shallow fully-connected stage (1 hidden layer), the most convolutional filters allowed (512 per layer), a modest number of dense units (16) and ReLU activation functions. For the training hyperparameters, small batch sizes were preferred (16), likely because this data is sufficiently regular to not require many samples to find a reasonable stochastic gradient; \textcolor{black}{also larger batch sizes can sometimes exceed GPU memory, resulting in a failed run}. Dropout was hardly used (0.001), which may be because the convolutional layers are already heavily regularized due to weight-sharing and the number of dense layers and their sizes are already constrained. There was little L1-regularization but some L2-regularization (0.001, 0.064), which again might be because it was better to constrain the number and sizes of hidden layers.

\section{Tactile pose estimation for a 3D edge}
\label{sec:6}

\begin{figure*}[t]
	\centering
	\begin{tabular}[b]{@{}c@{}}
		\textbf{\textcolor{black}{Hyperparameter optimization (3D edge)}} \\
		\includegraphics[width=2\columnwidth,trim={95 35 95 35},clip]{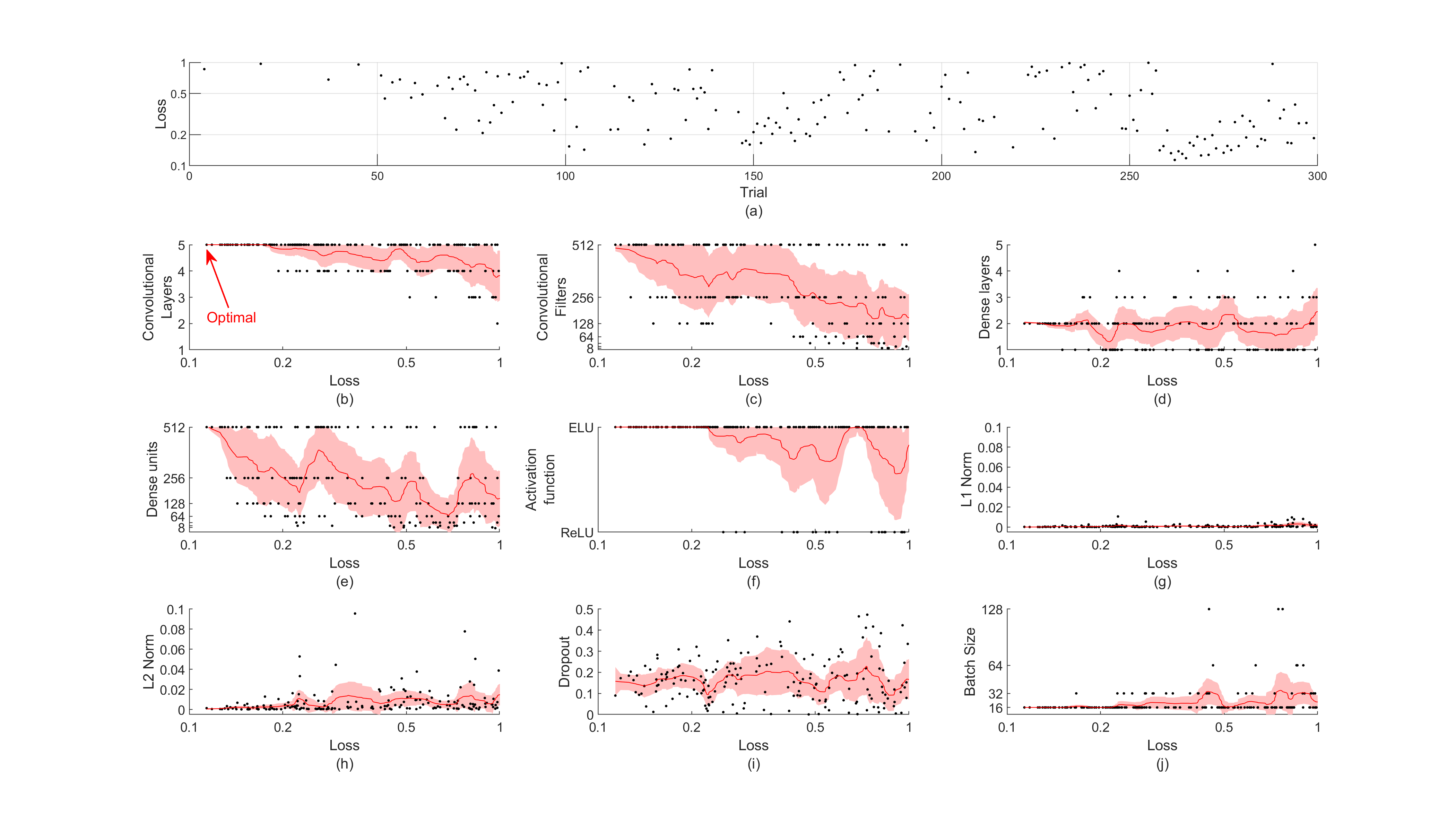}
	\end{tabular}
	\caption{\textcolor{black}{Hyperparameter optimization for the 3D edge. Top panels: loss against trial number. Bottom panel: hyperparameter dependence on loss.}}
	\label{fig10}
\end{figure*}

\subsection{3D edge model results}

The optimized poseNet performance was assessed for the 3D edge on a third dataset comprising 2000 tactile images labelled with horizontal pose, depth, roll, pitch and yaw. The data was again subject to unlabelled pose perturbations, with poses and perturbations sampled randomly (Table~\ref{tab2}).

Accurate performance was again obtained (Figure~\ref{fig9}), with MAEs between the predicted and labelled pose parameters of $0.6\,$mm, $0.2\,$mm, $1.4\,$deg, $1.6\,$deg and $6.4\,$deg. Clearly, these are less accurate than the surface, which reflects that pose prediction for the edge is a far more difficult regression problem. For example, the edge has 2 more pose parameters than the surface but the same amount of training data was used. Also, by nature of its geometry, the edge can feel similar over distinct poses: one example is that contacts at the horizontal extremes are ambiguous under yaw. These difficulties emphasise the importance of finding the best possible model fit using the hyperparameter optimization.

\subsection{Analysis of 3D edge model optimization}

For the 3D edge model, the hyperparameter optimization resulted in a 10-fold reduction of loss over the 300 training trials with the lowest losses after trial (Figure~\ref{fig10}, top panels). This improvement in loss was substantial but much less than for the 3D surface, which again reflects that estimating the pose of the edge is a more difficult problem. 

Losses were in the range 0.1--1 MSE, with the best losses occurring only near the end of the optimization. The convergence of the optimization can also be seen in scatter plots of the hyperparameters (Figure~\ref{fig10}), with the lowest losses bunched on the left of the plots. 

Overall, the optimization favoured network architectures (Table~\ref{tab3}) that have a deep convolutional stage (5 hidden layers), fairly shallow fully-connected stage (2 hidden layers), many convolutional filters (256 per layer), a moderate number of dense units (64) and ReLU activation functions. For the training hyperparameters, dropout is strongly preferred (0.2), and L1-regularization or L2-regularization rarely used ($\lesssim$0.001). Modest batch sizes were again preferred. The network and training hyperparameters are similar to those for the surface, apart from a greater use of dropout with the ReLU activation function. This appears to be a strategy for coping with the more difficult pose estimation problem for the edge. The different configuration emphasises the benefit of an automated process for model optimization.

\section{\textcolor{black}{Demonstration of 3D object exploration}}

Pose estimation is a fundamental capability of tactile sensing, because knowledge of relative pose enables a robot to control its interactions: local pose gives information on how to reposition the tactile sensor to maintain contact while moving safely over the object. To show pose estimation in action, the initial figure for this article displays trajectories generated by the PoseNet models in this paper applied to controlling a robot moving over a complex 3D surface and edge (Figure~\ref{fig1}).

{\color{black}We used a robot system comprising a BRL TacTip mounted on 6-DoF robot arm (IRB 120, ABB robotics). This robot was previously used to study 2D contour following~\cite{Lepora2019a}; we refer to that paper for details of the robot, software infrastructure and literature on tactile servoing.  

To demonstrate 3D surface following, we extended the previous 2D control policy to a 3D surface (3 pose variables) and a 3D edge (5 pose variables). This policy has two aims: (i)~move the sensor to remain normal to the object surface, and (ii) move the sensor tangentially along the surface (by 1\,mm per time step $t$). Here we implemented a proportional-integral (PI) controller in discrete time with output a change in 3D pose of the sensor in its reference frame
\begin{equation}\label{eq:9}\nonumber
\Delta\bm s(t) = K_{\rm p}\bm e(t) + K_{\rm i}\sum_{t'=0}^{t}\bm e(t').
\end{equation}
The gain matrices $K_{\rm p}$, $K_{\rm i}$ were diagonal with proportional gains 0.5 and integral gains 0.3 and 0.1 for translations and rotations respectively. The error $\bm e$ was evaluated between the predicted pose and a reference normal to the surface or edge. 

Using the PoseNet models (Section~\ref{sec:4}) for tactile pose estimation of a 3D surface (Section~\ref{sec:5}) and 3D edge (Section~\ref{sec:5}) gave successful object exploration over two complex 3D objects: surface exploration over a porcelein bust and edge following around a container top (Figure~\ref{fig1}). The exploration was over a region bounded by the robot reach and a joint-space singularity over the bust (on the right cheek). The estimated pose is shown in the insets, with surface normals in red and edge normals in blue along the trajectories.}

\section{Conclusions}

In this paper, we showed how optimal deep learning can give highly accurate models for robot touch using tactile images from optical tactile sensors. To illustrate this application, we considered local pose estimation of a 3D surface or edge in contact with the BRL tactile fingertip (TacTip), a biomimetic optical tactile sensor. To predict pose accurately, the trained network should be insensitive to how the sensor has reached its current pose. We developed models that are insensitive to motion-dependent shear by including representative examples of these motions as unlabelled perturbations of the training data. However, these made it difficult to predict pose, to the extent that improperly tuned models gave extremely sub-optimal results. For a systematic approach, we introduced Bayesian optimization of the network and training hyperparameters to find the most accurate models. \textcolor{black}{In consequence, the models were highly accurate ({\em e.g.} $0.1\,$mm depth, $0.3\deg$ surface orientation) even though the data were randomly perturbed by an unknown shearing motion. The models were also robust to surface shape and texture, as demonstrated by using the predicted poses to control a robot sliding the sensor over complex 3D objects (Figure~\ref{fig1}). }

\textcolor{black}{Overall, the approach for robot touch introduced here offers the potential for safe and precise physical interaction with complex environments, encompassing tasks from exploring natural objects to closed-loop dexterous manipulation. Even though we used the TacTip optical tactile sensor, a similar approach should apply to other high-accuracy tactile sensors such as the GelSight, provided they can slide repeatedly over objects without damage. This work aims to bring artificial tactile sensing one step closer to human performance, and so raises the question of whether humans use similar strategies during our own tactile interactions. In our view, soft tactile sensors like our fingertips cannot function usefully in natural environments unless they have a perceptual system with invariance to contact motion. As demonstrated here, appropriately trained deep neural networks can solve that problem.}

\section*{Acknowledgment} 

We thank NVIDIA Corporation for the donation of the Titan Xp GPU used for this research

\bibliographystyle{unsrt}
\bibliography{library}

\end{document}